\documentclass[runningheads]{llncs}
\usepackage{graphicx}

\usepackage{amsmath}
\usepackage{amssymb}
\usepackage{booktabs}
\usepackage{multirow}
\usepackage{tikz}
\usepackage{comment}
\usepackage{amsmath,amssymb} 
\usepackage{color}

\usepackage[accsupp]{axessibility}  

\usepackage[width=122mm,left=12mm,paperwidth=146mm,height=193mm,top=12mm,paperheight=217mm]{geometry}

\begin{document}
\pagestyle{headings}
\mainmatter
\def\ECCVSubNumber{100}  

\title{ReLER@ZJU Submission to the Ego4D Moment Queries Challenge 2022} 

\authorrunning{Shao et al.} 
\author{Jiayi Shao\textsuperscript{1} , Xiaohan Wang\textsuperscript{1} , Yi Yang\textsuperscript{1}} 
\institute{\textsuperscript{1} ReLER Lab, CCAI, Zhejiang University \\
shaojiayi1@zju.edu.cn, xiaohan.wang@zju.edu.cn, yangyics@zju.edu.cn}

\maketitle

\begin{abstract}
In this report, we present the ReLER@ZJU\textsuperscript{1} submission to the Ego4D Moment Queries Challenge in ECCV 2022. In this task, the goal is to retrieve and localize all instances of possible activities in egocentric videos. 
Ego4D dataset is challenging for the temporal action localization task as the temporal duration of the videos is quite long and each video contains multiple action instances with fine-grained action classes.
To address these problems, we utilize a multi-scale transformer to classify different action categories and predict the boundary of each instance. Moreover, in order to better capture the long-term temporal dependencies in the long videos, we propose a segment-level recurrence mechanism. Compared with directly feeding all video features to the transformer encoder, the proposed segment-level recurrence mechanism alleviates the optimization difficulties and achieves better performance. 
The final submission achieved Recall@1,tIoU=0.5 score of  37.24, average mAP score of 17.67 
 and took 3-rd place on the leaderboard.
\end{abstract}
\section{Introduction}
The goal of Ego4D MQ task\cite{grauman2022ego4d} is to retrieve and locate all possible instances in egocentric videos. Compared with traditional temporal action localization tasks(TAL) and benchmarks,  Ego4D MQ task is challenging with the following features: 1) Each video clip in the Ego4D dataset is extremely long, with an average duration of over 10 minutes. And each action instance also varies in duration ranging from about 1 second to several minutes. 2) Each video clip contains multiple action instances  which may be overlapped.  

In this case, we extract clip-level features and utilize a multi-scale transformer to better capture the temporal information of clips, fit action instances of different duration, and alleviate overlapping problems, inspired by the great success of transformer-based models\cite{zhang2022actionformer} in the TAL task. Meanwhile, as the duration of video clips is so long that directly feeding all clip features may lead to inefficient optimization and inferior performance, we utilize a segment-level recurrence mechanism, enabling the model to learn bidirectional contexts, motivated by \cite{dai2019transformer,yang2019xlnet}, making it possible to model long-term temporal dependency. The experimental results show that our method is effective for long videos and can achieve good performance on the validation and test set.

\section{Related work}
The MQ task can be formulated as the combination of action recognition or retrieval and action localization in the case of egocentric scenarios.  
Some previous works \cite{sudhakaran2019lsta,Wang_2021_ICCV,wang2020symbiotic} have gained great success in egocentric action recognition while multiple methods have also been proposed in TAL task, typically including actioness\cite{long2020learning}, anchor-free\cite{zhang2022actionformer} and anchor-based\cite{liu2020progressive} types. Inspired by the transformer for long-term dependency \cite{dai2019transformer,yang2019xlnet}, our idea is to exploit a more efficient way for long video temporal action localization.

\section{Methodology}
Denote a video clip dataset as $V=\{v_t\}_{t=1}^T$, where $T$ is the total RGB frames. Denote action instances ground-truth in each video as $\{B_m,E_m,C_m\}_{m=1}^{M_V}$, where $B_m$ and $E_m$ is the start time and end time, $C_m$ indicates the category of this instance and $M_v$ is the total number of action instances in video $V$. Our method is to predict action category $\tilde C_t$, boundaries $\tilde B_t, \tilde E_t$ of each location $t$ in the video utilizing a transformer-based model combined with segment-level recurrence mechanism.

\subsection{Base Model}
\begin{figure}[htbp]
\centering
\includegraphics[height=5.5cm]{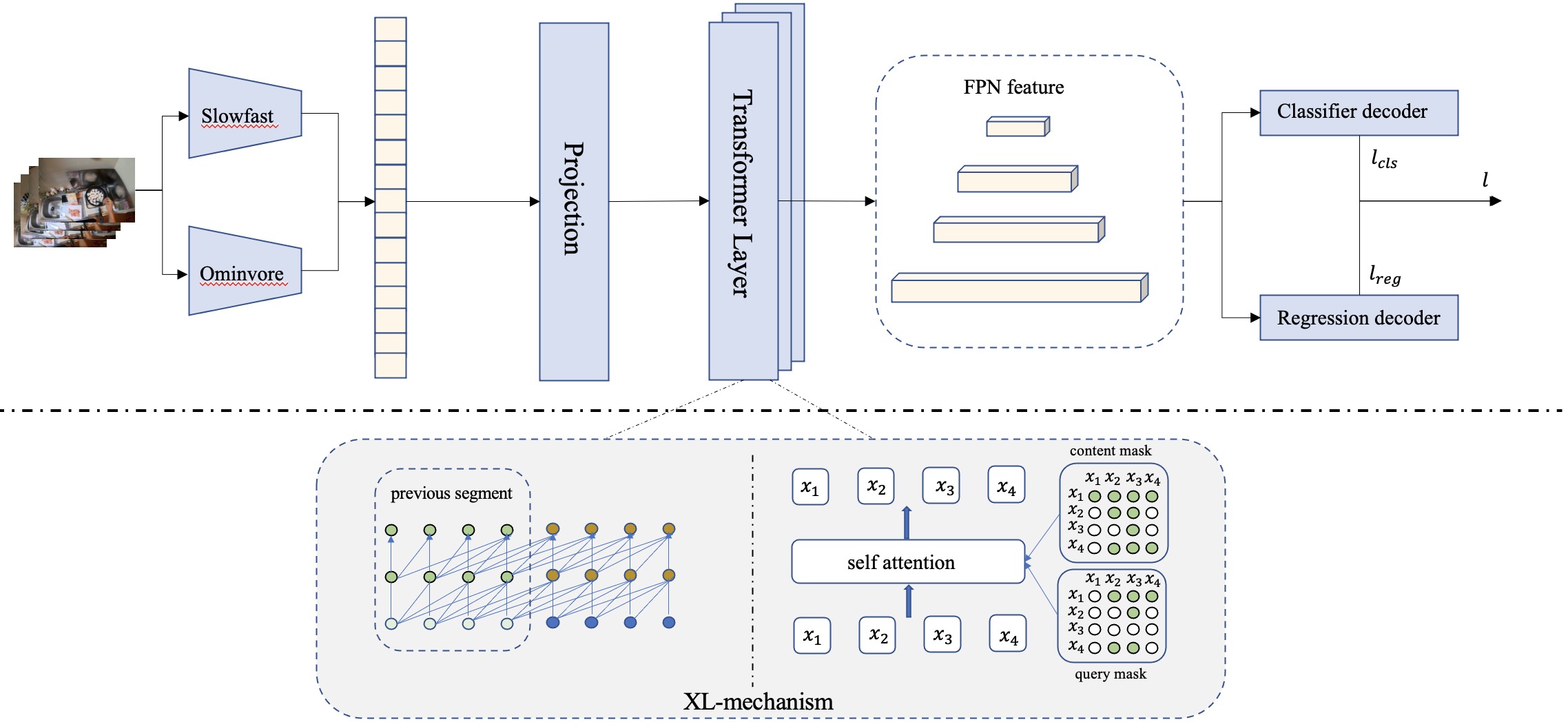}
\caption{The overall framework of our method}
\label{figure1}
\end{figure}

We develop our temporal action localization framework following the idea of Actionformer~\cite{zhang2022actionformer}.
As shown in Figure~\ref{figure1}, the input features are first projected into the latent space. Then a multi-scale transformer is used to get a better representation of the input video. We build our multi-scale mechanism adopting a FPN-based structure\cite{lin2017feature} that continuously downsamples each level feature using temporal convolution. After that two independent decoders are used to regress the boundary and predict the action category of each location respectively.  To optimize the model, we apply an IoU Loss for regression and a focal loss for classification, which can be formulated as:
\begin{equation}
    l=l_{cls}+l_{reg}=\frac{1}{N}\sum\limits_i l_{focal}(\tilde C_i, C_i) + \frac{1}{N}\sum\limits_i \mathbb{I}(C_i \geq 1) (1-\frac{|\tilde \varphi_i \cap \varphi_i|}{|\tilde \varphi_i \cup \varphi_i|})
\end{equation}
where $\varphi_i=(B_i,E_i)$. In the inference phase, we treat each frame-level prediction $\{\tilde B_t, \tilde E_t, \tilde C_t\}$ as a temporal bounding box and apply Soft-NMS\cite{bodla2017soft} to suppress redundancy. It's worth noting that because of dense annotations in Ego4D videos, we keep a relatively large number of predictions, i.e., 1000, for final results.
We use the provided slowfast features\cite{fan2020pyslowfast} as well as Omnivore features\cite{girdhar2022omnivore} as input. where these two models all use a window size of 32 and a stride of 16 to extract video-level features.

\subsection{Segment-level Recurrence Mechanism}
We utilize the segment-level recurrence mechanism to boost the long-temporal modeling and relieve the optimization difficulties. Specifically, a long video is first split into several segments. Each segment forwards into an attention mechanism to get the local representation of segments. 
Meanwhile, to avoid context fragmentation and learn a better global representation, attention also conducts between segments: the hidden state output of the previous segment is cached and reused in the next new segment's computation. Moreover, we adopt an XL-based\cite{yang2019xlnet} mechanism to boost the encoder, where an input token can not only see its previous but also future tokens via designing and shuffling the content and query mask. For example, as shown in Figure~\ref{figure1} below, this content and query mask indicates the actual input sequence is $x_3 \rightarrow x_2 \rightarrow x_4 \rightarrow x_1$ and $x_2,x_3,x_4$ all contribute to  $x_1$'s representation.

\section{Experiments}
In all experiments, we use the provided slowfast\cite{fan2020pyslowfast} and omnivore\cite{girdhar2022omnivore} clip features and simply concatenate them channel-wise as input. We upsample the input sequence to 1024, choose 8 fpn levels, use a learning rate of $1e^{-3}$ with a weight-decay of $0.1$ and train $15$ epochs.

For the action retrieval and localization task, as shown in Table~\ref{table 1}, Our ensemble model  achieves the 3-rd best performance on both Recall@1x-tIoU=0.5 and average mAP metrics  among all challenge entries.   

\begin{table}[htbp]
\centering
\caption{Leaderboard results on the test set of Ego4D MQ challenge}
\label{table 1}
\scalebox{0.8}{
\begin{tabular}{lccccccccccccc} \toprule

Method/Entries & Recall@1x, tIoU=0.5 & Average mAP    \\ \midrule
ohohoh          & 41.13               & 23.59          \\ 
jenny-jenny     & 42.54               & 21.76         \\ 
\textbf{ours}   & \textbf{37.24}            & \textbf{17.76} \\ 
TempSeg         & 27.54              & 11.33          \\ 
icego           & 29.78               & 11.00          \\ 
ego vlp         & 28.03               & 10.33          \\ 
mtz             & 13.80              & 6.51          \\ 
baseline(VSGN\cite{zhao2021video})  & 24.25               & 5.68           \\ \bottomrule
\end{tabular}
}
\end{table}

\begin{table}[htbp]
\centering
\caption{Ablation results on the val set of Ego4D MQ challenge}
\label{table 2}
\resizebox{\columnwidth}{!}{%
\begin{tabular}{lcccccccc}
\toprule
Method           & mAP tIoU=0.1 & mAP tIoU=0.3 & mAP tIoU=0.5 & avg mAP & Recall@1 tIoU=0.5 & mem /GB \\ \midrule
base             & 18.74        & 13.72        & 9.44         & 13.92       & 33.19              & 10.9    \\ 
split,length=128 & 18.95        & 13.82        & 9.07         & 13.95       & 32.91              & 8.7     \\ 
split,length=256 & 18.56        & 13.90        & 9.68         & 14.03       & 33.70              & 8.9     \\ 
base+XL          & 19.69        & 14.62        & 10.23        & 14.84       & 35.27              & 11.1    \\ \bottomrule
\end{tabular}%
}
\end{table}

\textbf{Ablation Study:} Directly modeling the long-term dependencies by feeding all video clips to the transformer encoder may hinder optimization. We investigate two mechanisms to alleviate this problem. As shown in Table~\ref{table 2}, \textbf{Base} means the whole input sequence forwards into attention without any other processing, \textbf{Split, length=x} means the input sequence is split into segments with a length of $x$, \textbf{XL} means using segment-level recurrence mechanism. To better exploit xl-mechanism, we only use it in the first level of FPN to replace the transformer layer as the input feature sequence is the longest. The results indicate the effectiveness of our method for this particular task.

\begin{figure}[htbp]
\centering
\includegraphics[height=4.5cm]{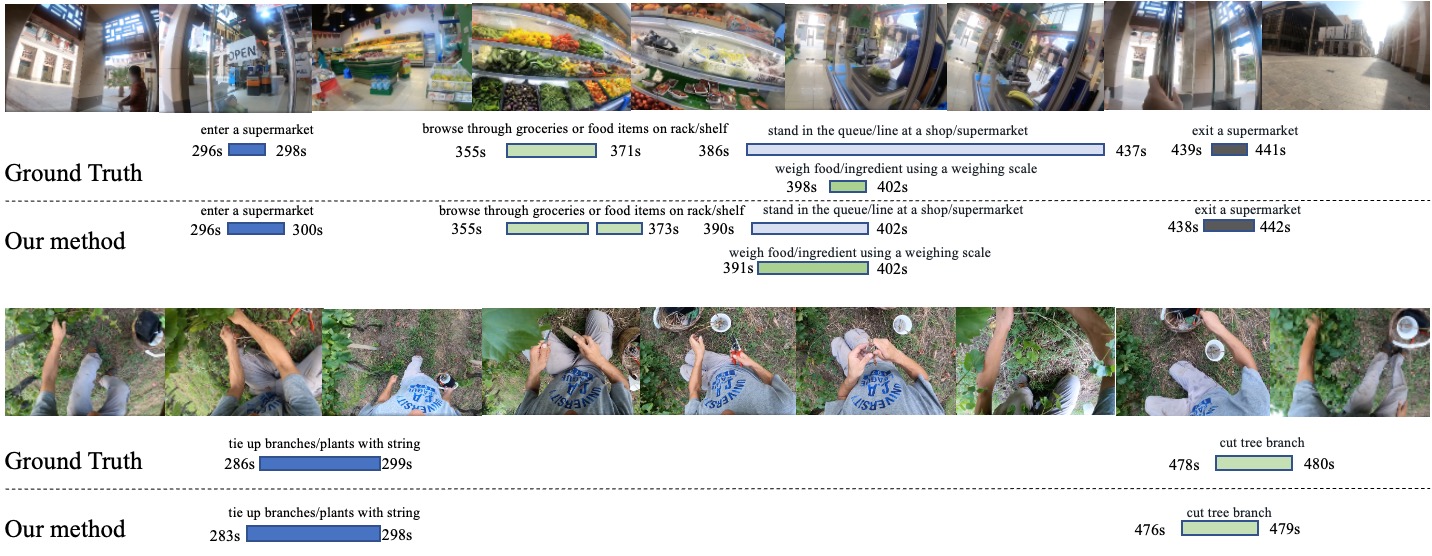}
\caption{The positive and negative examples of our method}
\label{figure2}
\end{figure}

\textbf{Qualitative Results:}Figure~\ref{figure2} shows several videos in the validation set of the MQ challenge, including positive and negative results of our method. Specifically, our model performs relatively not bad for normal actions but still suffers from wrong predictions of the very short or long actions. For short actions e.g. \emph{cutting tree branch}, they may harm the performance since they are more sensitive to predicted boundaries. While for long actions, e.g. \emph{browse through groceries or food items on rack/shelf}, our model may get stuck in outputting several fragmentized temporal action proposals rather than one long temporal prediction.

\clearpage
%
%
\bibliographystyle{splncs04}
\bibliography{cit}
\end{document}